\def\year{2019}\relax
\documentclass[letterpaper]{article} 
\usepackage{aaai19}  
\usepackage{times}  
\usepackage{helvet}  
\usepackage{courier}  
\usepackage{url}  
\usepackage{graphicx}  
\usepackage{subcaption}
\usepackage{amsmath}
\usepackage{cleveref}
\begin{document}
%
\title{Multi-shot Person Re-identification through Set Distance with\\ Visual Distributional Representation}
\author{Ting-Yao Hu, Xiaojun Chang, and Alexander G. Hauptmann\\
Language Technologies Institute, Carnegie Mellon University\\
5000 Forbes Avenue\\
Pittsburgh, Pennsylvania 15213\\
}
\maketitle
\begin{abstract}

Person re-identification aims to identify a specific person at distinct times and locations.
It is challenging because of occlusion, illumination, and viewpoint change in camera views.
Recently, multi-shot person re-id task receives more attention since it is closer to real-world application. 
A key point of a good algorithm for multi-shot person re-id is the temporal aggregation of the person appearance features.
While most of the current approaches apply pooling strategies and obtain a fixed-size vector representation, these may lose the matching evidence between examples.
In this work, we propose the idea of \textit{visual distributional representation}, which interprets an image set as samples drawn from an unknown distribution in appearance feature space.
Based on the supervision signals from a downstream task of interest, the method reshapes the appearance feature space and further learns the unknown distribution of each image set.
In the context of multi-shot person re-id, we apply this novel concept along with Wasserstein distance and learn a distributional set distance function between two image sets.
In this way, the proper alignment between two image sets can be discovered naturally in a non-parametric manner. 
Our experiment results on two public datasets show the advantages of our proposed method compared to other state-of-the-art approaches.

\end{abstract}

\section{Introduction}
Person re-identification (person re-id) aims to identify a specific person at distinct times and locations. 
It is an essential task for several applications, such as long-term person tracking across camera views \cite{Ristani_2018_CVPR}. 
Re-id task is still challenging due to the appearance variations of person. 
These variations usually come from occlusion, illumination, and viewpoint change in camera views.
Recent approaches usually treat re-id task as a retrieval problem: given a query based on a single image or a set of images, and a gallery set of candidate person images, we need to rank these candidates according to some similarity metrics. 
Researchers have considered two scenarios, single-shot and multi-shot, for person re-id task. 
A lot of previous works have focused mainly on the single-shot scenario, while only a few lie in the latter. 
However, multi-shot person re-id is more suitable for practical surveillance applications, since person tracklets are available by applying object detection and tracking algorithms. 
In this paper, we investigate person re-id task in the multi-shot scenario.

\begin{figure}
  \centering
    \includegraphics[width=0.4\textwidth]{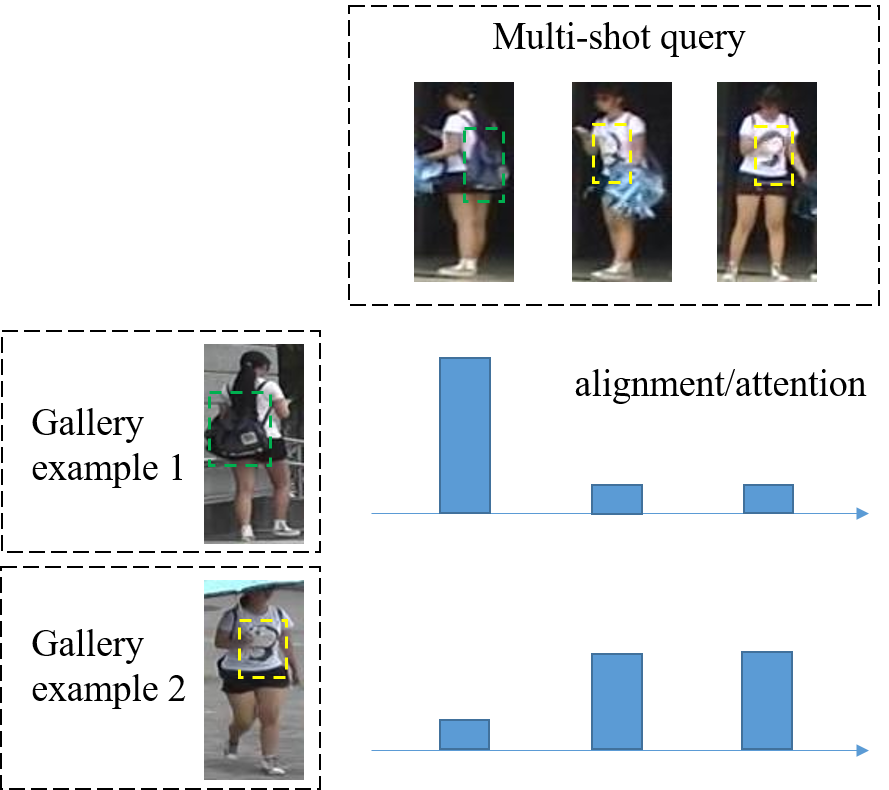}
  \caption{In multi-shot re-id task, a good algorithm is supposed to focus on different images of the query while comparing to different examples in the gallery set. Our algorithm aims to discover the most proper alignment between the query and gallery examples.}
  \label{fig:reid1}
\end{figure}

Multi-shot person re-id methods require the comparison between two sets of images. 
Thanks to the rapid development of deep learning techniques, recent  approaches adopt convolutional neural network (CNN) to extract the appearance feature of each image, and take temporal pooling strategies to aggregate an appearance feature sequence and form a fixed-size vector representation. 
Common temporal pooling strategies include mean and max pooling, recurrent neural network (RNN), and attention models. 
Finally, the dissimilarity between two vectors is calculated based on a distance function, such as the Euclidean and cosine distance. 
However, using a fixed-size vector as the representation of a set of images, previous re-id algorithms may ignore the matching evidence between two person tracklets.
For example, when the system compares the image set of identity $a$ to the image sets of $a'$ and $b$ from the gallery set, the evidence showing that $a$ and $a'$ are identical, and that showing $a$ and $b$ are not, may come from different images in set $a$.
Hence, instead of aggregating appearance features by conventional pooling strategies, a better way should be to represent an image set by the whole appearance feature set, and discover the alignment/attention between two sets, as shown in \Cref{fig:reid1}.

In this paper, we propose the idea of \textit{visual distributional representation} 
and use it to solve multi-shot person re-id task. 
Our approach treats a set of images, or a person tracklet as samples drawn from a probability distribution in appearance feature space.
Based on this concept, our multi-shot person re-id algorithm consists of three parts: an appearance feature extractor, a probability estimator, and a distributional distance function.
Specifically, we choose the \textit{Wasserstein distance} between distributions as the function calculating the dissimilarity between two image sets. 
By doing so, we can model the diversity and uncertainty of each image set, and embed the alignment/attention mechanism into our re-id algorithm without introducing additional parameters.
Furthermore, the dissimilarity score also serves as a supervision signal to reshape the appearance feature extractor and learn the distributions of all the image sets.

We conduct experiments on MARS \cite{zheng2016mars} and iLIDS-VID \cite{ilidsvid} datasets. The results demonstrate that the proposed person re-id algorithm based on visual distributional representation outperforms some conventional pooling strategies, and achieve state-of-the-art performance.

In summary, the contributions of this work are three folds: 
(1) We outline the concept of visual distributional representation, which aims to preserve the diversity and uncertainty of a set of observations. 
(2) Based on this concept, we utilize Wasserstein distance to design an image set distance function and build a network architecture to solve multi-shot person re-id problem. This architecture is simple yet effective in the alignment between two image sets.  
(3) We show that our proposed method outperforms several baselines and provides competitive/superior performance comparing to state-of-the-art approaches.

\section{Related Work}
Person re-identification task has been investigated for a long time. Earlier works in this area can be categorized into two major parts: feature extraction and metric learning. For feature extraction, a variety of visual features have been proposed to capture the appearance of a person, such as color histogram \cite{colorhist}, Local Binary Pattern (LBP) \cite{lbp}, and Local Maximal Occurrence (LOMO) \cite{lomo}. Some other works leveraged metric learning techniques to discover a discriminative distance measure in appearance feature space, such as KISSME \cite{kissme}, LFDA \cite{lfda}, and LMNN \cite{lmnn}.
On the other hand, deep learning based methods \cite{deep2,ding2015deep,ahmed2015improved,suh2018part} have been proposed to reduce the demand of hand-crafted feature design, and jointly conduct feature extraction and metric learning.

Recently, more and more works focused on multi-shot person re-id task, since it is closer to the requirement of a real system.
The re-id algorithms receive a person tracklet as a input and should take the advantage of the availability of multi-shot images. 
\cite{mclaughlin2016recurrent} proposed an RNN model to encode temporal information, and adopted mean/max pooling to aggregate features over each dimension. 
\cite{jointly} built a network incorporating spatial pyramid attention and temporal RNN jointly. 
In \cite{mclaughlin2016recurrent} and \cite{jointly}, the authors utilized both RGB and optical flow channel as network input. 
Most of the recent work considered RGB channels to be the input, and focused on the development of attention models. 
\cite{liu2017quality} used temporal attention to estimate the quality score of each image, and re-weighted the appearance features according to quality scores. 
\cite{li2018diversity} proposed an attention network that automatically discovered a diverse set of discriminative body parts. 
The features from these local body regions were aggregated by temporal attention.
\cite{zhou2017seeforest} used RNN to model spatial and temporal attention simultaneously.
\cite{si2018dual} proposed a dual attention model to perform context-aware feature sequence comparison.

Wasserstein distance has been successfully applied to many different applications. \cite{arjovsky2017wasserstein} improve the training process of generative adversarial network (GAN). 
\cite{frogner2015learning} utilized it to design a new loss function for multi-label classification. 
\cite{shen2018wasserstein} measured the distance between source and target domains in domain adaptation task.
Another set of works \cite{cuturi2013sinkhorn,ye2016simulated} focused on reducing the computation cost of Wasserstein distance calculation, which is crucial to the inference process of our framework.

\section{Proposed Method:\\ Visual Distributional Representation}
We propose the concept of visual distributional representation, and apply it to solve multi-shot person re-id task. 
Our algorithm learns how to represent an image set as a probability distribution, and conducts set-to-set comparison by calculating the distance between two distributions.
We claim that our method can effectively preserve the diversity of person appearance, and discover a proper alignment between two image sets.
In the mean time, it does not increase the number of parameters comparing to models with naive mean pooling.
In this section, we will elaborate on the details.

\subsection{Problem Definition: Multi-shot Person Re-ID}
In multi-shot person re-id task, multiple images of one identity are available. 
A key point of a good algorithm is its set-to-set comparison. 
Given two sets of images $S_i = \{I^i_1, I^i_2,...,I^i_{n_i}\}$ and $S_j = \{I^j_1, I^j_2,...,I^j_{n_j}\}$, we have to define a proper set distance function $D(S_i,S_j)$ to calculate the dissimilarity between them. 
In most of the previous works, this function can be decomposed into three parts: an appearance feature extractor, a pooling method, and a distance metric in feature vector space. 
The appearance feature extractor is usually implemented as a CNN. And the pooling method aggregates the appearance features into a fixed-size vector. 
For example, if we adopt mean pooling and the Euclidean distance, the set distance function would become:
\begin{equation}
D(S_i,S_j) = ||\frac{1}{n_i}\sum_{k=1}^{n_i} f(I^i_k) -  \frac{1}{n_j}\sum_{k=1}^{n_j} f(I^j_k)||
\label{mean_pooling}
\end{equation}
where $f$ represents the appearance feature extractor.
Other possible choices of pooling method include max pooling, RNN and attention models.

However, the conventional set distance functions like \Cref{mean_pooling} fail to align the evidence appearing in two sets, and thus deteriorate the performance. 
This is because previous pooling methods aggregate appearance features and generate a fixed-size vector before doing the comparison between two image sets.
To resolve this issue, we propose the concept of visual distributional representation, which naturally handles the diversity and uncertainty in a set of images.

\subsection{Visual Distributional Representation}
We assume that an image set can be treated as samples of a probability distribution in appearance feature space. 
The purpose of visual distributional representation is to find a proper distribution to describe a image set, and optimize the function for appearance feature extraction, simultaneously. Hence, the basic components of visual distributional representation learning framework include an appearance feature extractor $f$, a probability estimation method $pe$, and a downstream task, which provides the supervision signal for the optimization of $f$.

In the context of multi-shot person re-id task, we use visual distributional representation to describe the appearance of a person, and design a new type of set distance function, which can be formulated as followed:
\begin{equation}
\begin{split}
D(S_i, S_j) = &d(\nu_i, \nu_j) \\
\nu_i = &pe(f(I^i_1), f(I^i_2),...,f(I^i_{n_i}))\\
\end{split}
\end{equation}
where $\nu_i$ and $\nu_j$ are the visual distributional representations of $S_i$ and $S_j$, respectively, and $d$ is a distance measure between two distributions. It is important to define all three basic components: $f$, $pe$, and $d$ properly. In this study, we mainly focus on the investigation of probability estimator $pe$ and distributional distance measure $d$.

The goal of the probability estimator $pe$ is to summarize a set of appearance features by estimating its distribution. It is in parallel to the pooling strategies in conventional person re-id approaches.
In this paper, we consider two types of probability estimator:
(1) kernel density estimation (KDE) with Dirac delta function $\delta$: $\hat{\nu_i} = \frac{1}{n_i}\sum_{k=1}^{n_i} \delta (f(I^i_k))$
(2) Gaussian estimation (GE): assuming that the samples in appearance feature space follow a multi-variate Gaussian distribution $\hat{\nu_i} \sim N(m_i, \Sigma_i)$, while mean $m_i$ and covariance $\Sigma_i$ can be obtained from the appearance features easily.

Since each person tracklet is represented as a distribution, we need a distributional distance measure to calculate the dissimilarity between two tracklets. In this study, we specifically choose Wasserstein distance because it discovers the optimal alignment between two tracklets. More details are discussed in the following.

\begin{figure}
  \centering
    \includegraphics[width=0.5\textwidth]{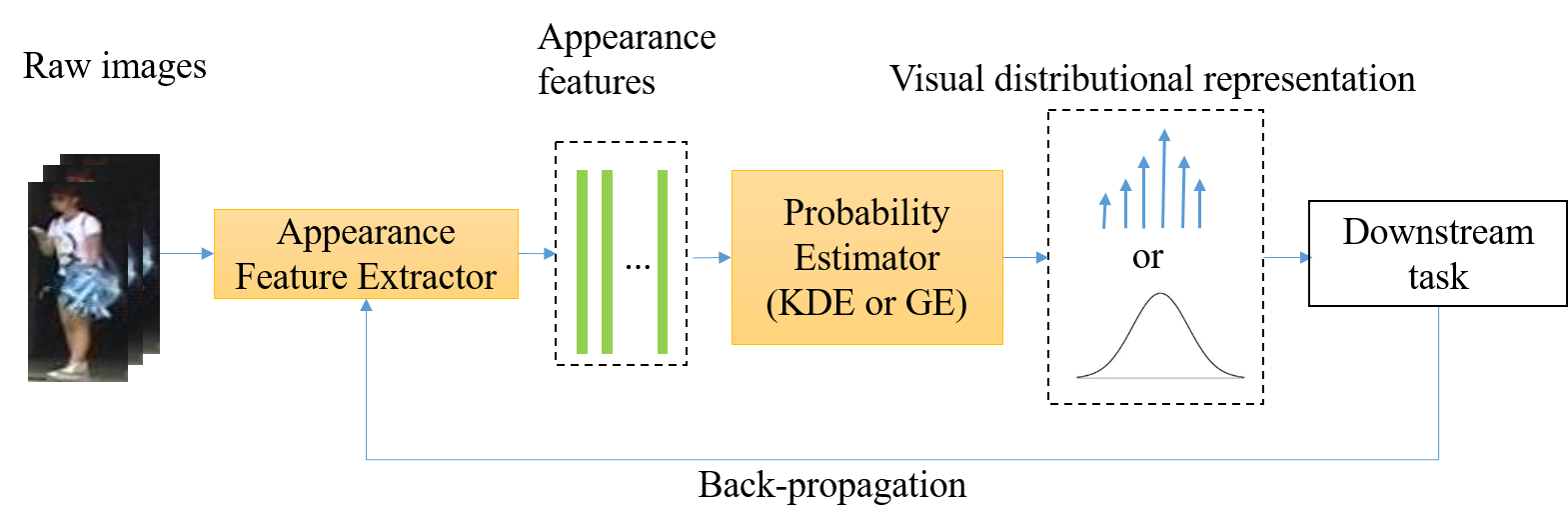}
  \caption{The whole process of visual distributional representation learning.
}
  \label{fig:reid_dr}
\end{figure}

To sum up, visual distributional representation is proposed to replace the conventional pooling strategies. Comparing to a fixed-size vector representation, it effectively captures the diversity and uncertainty of a set of images.
\Cref{fig:reid_dr} indicates the whole process of learning framework for visual distributional representation

\subsection{Visual Distributional Representation Learning with Wasserstein Distance}
We design our multi-shot person re-id algorithm based on the concept of visual distributional representation with Wasserstein distance.
In this section, we first revisit the basic idea of Wasserstein distance, discuss practical issues of visual distributional representation learning, and describe the details of our model.

\subsubsection{Wasserstein Distance Revisit}
Given a metric space $(\chi, d_{\chi})$, where $d_{\chi}(x,y)$ is a distance function between two elements $ x \in \chi$ and $ y \in \chi$, the p-th Wasserstein distance between two distributions $\nu_x$ and $\nu_y$ can be defined as:
\begin{equation}
    W^p(\nu_x,\nu_y)=\Big( \inf_{\pi \in \Pi (\nu_x,\nu_y)} \int_{\chi \times \chi} d_{\chi} (x,y) d\pi(x,y) \Big)^{\frac{1}{p}}
\end{equation}
where $\Pi (\nu_x,\nu_y)$ is the set of all joint probability $\pi$ on $\chi \times \chi$ with marginals $\nu_x$ and $\nu_y$. In this study, we choose to set $p=2$ so $d_{\chi}$ is the Euclidean distance.

In the discrete case, we need to compare two empirical measures $\hat{\nu_x} = \frac{1}{n_x}\sum_{k=1}^{n_x} \delta(x^{(k)})$ and $\hat{\nu_y} = \frac{1}{n_y}\sum_{l=1}^{n_y} \delta(y^{(l)})$ represented by the uniformly weighted sums of $n_x$ and $n_y$ Diracs delta with mass at positions $x^{(k)}$ and $y^{(l)}$.
$\Pi (\hat{\nu_x}, \hat{\nu_y}) = \{ \sum_{k,l} w_{k,l} \delta(x^{(k)}) \delta(y^{(l)}) | \sum_k  w_{k,l}=\frac{1}{n_y},  \sum_l  w_{k,l}=\frac{1}{n_x}\}$ is the valid set of joint probability distributions, whose element can be noted as a $n_x \times n_y$ matrix $P$. 
Hence, 2-Wasserstein distance becomes:
\begin{equation}
W^2(\hat{\nu_x}, \hat{\nu_y}) = \min_{P \in \Pi (\hat{\nu_i}, \hat{\nu_j})} \big\langle P, M \big\rangle_F
\label{OP}
\end{equation}
where $M$ is a $n_x \times n_y$ distance matrix and $M_{kl}$ stores the Euclidean distance between $x^{(k)}$ and $y^{(l)}$.
Finding the discrete version Wasserstein distance is usually known as optimal transport problem \cite{villani2008optimal}.

The discrete version of Wasserstein distance provides a natural way to conduct set-to-set comparison. 
For each element in a set, it considers only the distances with the nearest neighbors in the other set. 
When the appearance evidence of re-id is not dense, this approach helps the system automatically discover a proper alignment, and makes set-to-set comparison more robust.
In our visual distributional representation learning framework, if we use kernel density estimation (KDE) with Dirac delta function to represent each person tracklet, then the final set distance function would become:
\begin{equation}
D_E(S_i,S_j) = W^2(\hat{\nu_i}, \hat{\nu_j})
\label{Exact}
\end{equation}
where $\hat{\nu_i} = \frac{1}{n_i}\sum_{k=1}^{n_i} \delta (f(I^i_k))$ and $\hat{\nu_j} = \frac{1}{n_j}\sum_{k=1}^{n_j} \delta (f(I^j_k))$.

\subsubsection{Approximation of Wasserstein Distance}
The efficiency of Wasserstein distance calculation is an important issue to our proposed model.
Finding the exact solution of \Cref{Exact} is equivalent to solving a linear programming problem, which costs $O(n^3 log(n)), n = \max(n_i,n_j)$ \cite{pele2009fast}. 
In the training phase, we randomly sample a small number of images for each person and avoid solving a large-scale problem. 
In the testing phase, however, we need to process all the images of each person in order to achieve the best performance. 
Practically, the number of images of each person, especially in the gallery set, could be very large. 
Hence, to make the inference process efficient, we adopt two types of approximation.

The first approach is to smooth the original Wasserstein distance with an entropy regularization term \cite{cuturi2013sinkhorn}:
\begin{equation}
    D_S(S_i,S_j) = \min_{P \in \Pi (\hat{\nu_i}, \hat{\nu_j})}  \big\langle P, M \big\rangle_F - \frac{1}{\lambda} h(P)
\label{WSI}
\end{equation}
where $h(P) = \sum_{k,l} P_{k,l} log (P_{k,l})$ is the entropy of joint probability. It can be solved by iterating Sinkhorn’s update \cite{cuturi2013sinkhorn}, which costs only $O(n^2)$.

Entropy regularized Wasserstein distance is also called Sinkhorn distance. It still takes the distances between all the image pairs into account, but gives more weight to nearest neighbors. When $\lambda \to \infty$, it is equivalent to the original Wasserstein distance.

The second approach is to make a parametric assumption: the extracted appearance features of an image set follow a Gaussian distribution. In the case of comparing two Gaussian distributions $\hat{\mu_i} \sim N(m_i, \Sigma_i)$ and $\hat{\mu_j} \sim N(m_j, \Sigma_j)$ , Wasserstein distance can be calculated by the following \cite{dowson1982frechet}:
\begin{equation}
\begin{split}
    &D_G(S_i,S_j) =
    W^2(\hat{\mu_i}, \hat{\mu_j}) =\\ &||m_i-m_j||^2+Tr||\Sigma_i+\Sigma_j-2(\Sigma_i^{1/2}\Sigma_j\Sigma_i^{1/2})^{1/2}||
\end{split}
\label{WG}
\end{equation}
$m_{i}$, $m_{j}$, $\Sigma_{i}$ and $\Sigma_{j}$ can be estimated directly by the mean and covariance of appearance features of $S_i$ and $S_j$. 
Comparing to entropy regularization, this approach does not require accessing the Euclidean distance matrix $M$, so it is much more efficient. 
Obviously, to use this approximation in our framework, we also need to choose the probability estimator with Gaussian assumption (GE).

\begin{figure}
  \captionsetup[subfigure]{justification=centering}
  \begin{subfigure}{0.5\textwidth}
  	\centering
    \includegraphics[width=0.85\textwidth]{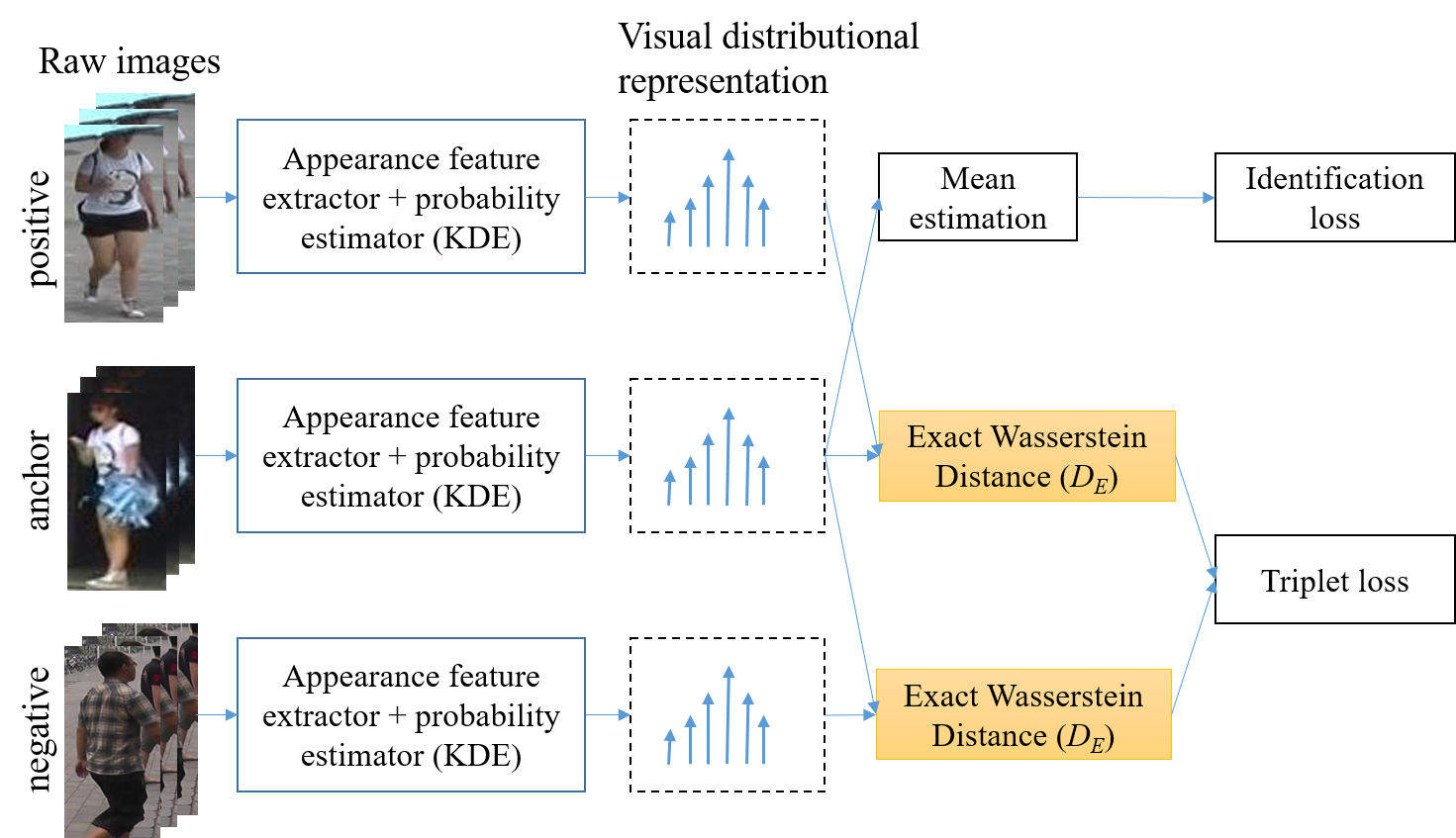}
    \caption{Training phase}
  \end{subfigure}\vspace{1em}
  \begin{subfigure}{0.5\textwidth}
  	\centering
    \includegraphics[width=0.8\textwidth]{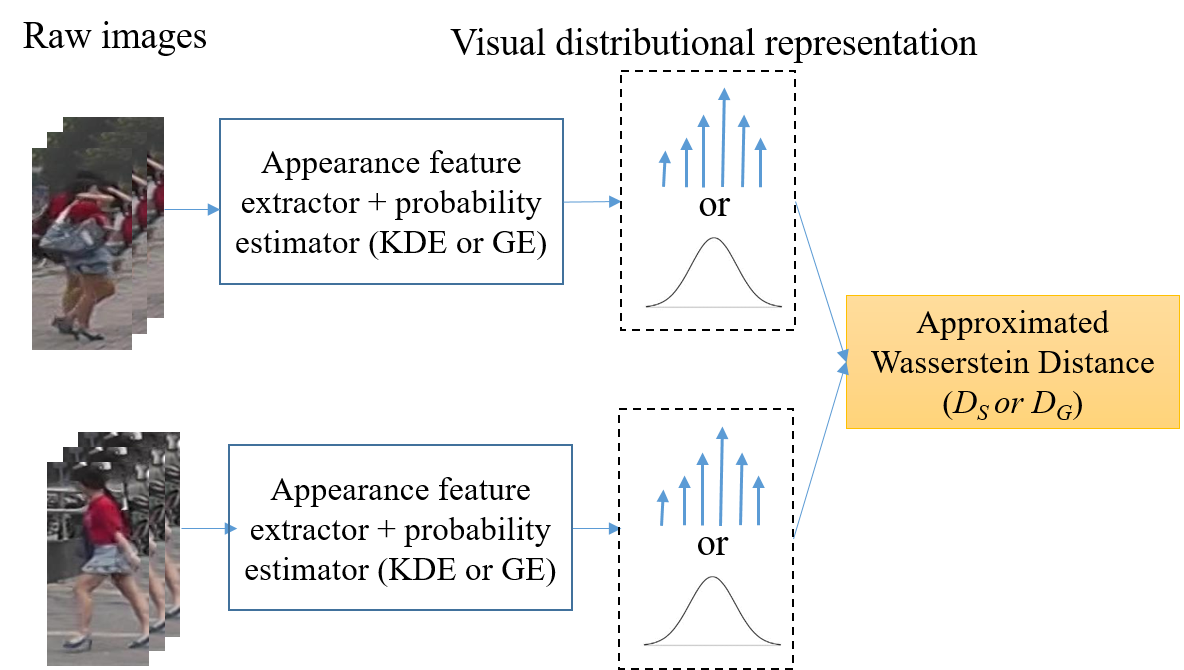}
    \caption{Testing phase}
  \end{subfigure}
  \caption{The architecture of our model in the training and testing phases}
  \label{fig:reid2}
\end{figure}

\subsection{Model Details}
\Cref{fig:reid2} illustrates the architecture of our model on the strength of visual distributional representation learning.
In the training phase, our model extracts appearance features, estimates the visual distributional representation by KDE with Dirac delta, and optimize the parameters in appearance feature extractor based on training objective function. The training objective is defined with exact 2-Wasserstein distance (\Cref{Exact}).
In the testing phase, given a person tracklet as query, our model calculates the dissimilarity between the input tracklet and all tracklets in gallery set. The dissimilarity can be determined by two types of Wasserstein distance approximation, \Cref{WSI} and \Cref{WG}, which require the distributions estimated by KDE with Dirac delta and GE, respectively. To eliminate the influence of outlier images in person tracklet, we apply a moving average filter to appearance feature sequences in the testing phase:
\begin{equation}
f'^{(k)}_i = \frac{1}{K} \sum_{l=1}^{K} f(I_{k+l}^i)
\label{MA}
\end{equation}
and then estimate the distribution of this tracklet on smoothed appearance features $f'^{(k)}_i$.

\subsubsection{Training Objective Function}
The overall objective function in the training phase consists of two parts: triplet loss and identification loss. 

Let $(S_a, S_p, S_n)$ be a triplet input, where $S_a$, $S_p$ and $S_n$ are the anchor, positive, and negative examples, respectively. Anchor and positive examples belong to the same identity, while the negative example is from a different one.
Triplet loss \cite{triplet} forces the distance between the positive pair $D_E(S_a, S_p)$ to be smaller than the distance between the negative pair $D_E(S_a, S_n)$ with a margin $\Delta$:
\begin{equation}
    L_{triplet} = \sum_{a=1}^{B} max(0, D_E(S_a, S_p)-D_E(S_a, S_n)+\Delta)
\end{equation}
where $B$ is the batch size. In practice, we apply batch-wise hard-negative mining \cite{hermans2017defense}, selecting the most distant positive example and most similar negative example within a batch for each anchor.

The identification loss aims to categorize each set of images to the correct identity:
\begin{equation}
L_{ID} = \sum_{a=1}^{B}\sum_{k=1}^{N_{id}} q_{ak} log( p(k|F(\nu_a)) )
\end{equation}
where $N_{id}$ is the total number of identities, and $\nu_a$ is the visual distributional representation of input image set $S_a$.
$q_{ak}=1,0$ indicates if $S_a$ belongs to $k$-th identity.
$log (p(k|F(S_a)))$ is obtained from a logistic regression layer taking $F(\nu_a)$, the mean estimation of $\nu_a$, as input, which is equivalent to the output of mean pooling strategy. 

The total loss is the combination of the two: $L_{total} = L_{triplet}+L_{ID}$. By observing the log during the training phase, we find that the network can usually decrease triplet loss to zero, so we don't need to introduce a hyper-parameter to adjust the balance between two loss functions. 
Based on this total loss, the whole network is trained in an end-to-end manner using back-propagation.

\subsection{Discussion}
Attention mechanism \cite{RQEN,li2018diversity,jointly,liu2017quality} is another common approach trying to improve naive pooling methods like mean and max pooling.
An additional branch of network is applied to calculate the weight of each image in the set, and assess the final representation by leveraging the weighted summation of the extracted features of the whole set. 
This type of methods can effectively deal with the noisy elements in  an image set. However, the aggregation process of final representation is still independent of other identities in the training data. Thus, it is not likely to capture the sparse evidence for set-to-set comparison.
There were also several recent approaches dealing with set-to-set comparison \cite{zhou2017seeforest,si2018dual} directly. They both introduced dual attention frameworks, performing image pair alignment. 
However, these frameworks use additional parameters to model the attention mechanism, and increase the risk of over-fitting. 
In our model, Wasserstein distance naturally incorporates feature alignment, and no additional parameters are required.

In many applications, Wasserstein distance estimation is done by Kantorovich-Rubinstein dual form \cite{kantorovich1958space}. It requires finding the solution of an optimization problem in 1-Lipschitz function space, which is often implemented by another deep neural network \cite{arjovsky2017wasserstein}. 
This dual form solution is not applicable to our case. 
We need to estimate the distance between each pair of persons. 
For each person tracklet, only several hundreds of examples are available at most, which is not enough to train a network.

While we choose Wasserstein distance in this work, our learning framework accepts other types of distance measure between distributions, such as total variance and symmetric KL divergence. 
One can investigate the advantages of these choices, although it is not the main focus of this work.

\section{Experiments}
In this section, we evaluate our proposed model for multi-shot person re-id on two public datasets: iLIDS-VID \cite{ilidsvid} and MARS \cite{zheng2016mars}.  

\subsection{Datasets and Evaluation Protocol}
The iLIDS-VID dataset contains 300 identities. For each identity, 2 tracklets are captured from two non-overlapping cameras, respectively. The number of frames of each tracklet ranges from 23 to 192, with an average number of 73. The bounding boxes of trackelts are annotated by humans.
The MARS dataset contains 1,261 identities and 20,715 tracklets. 
The bounding boxes of tracklets are detected and tracked by DPM detector and GMMCP tracker, respectively.
There are 6 camera views, and each identity is captured by at least two views. 
3,248 distractor tracklets appeared due to false detection or tracking.

Following the standard evaluation protocol of iLIDS-VID, we randomly split 50\% of identities for training and 50\% for testing, and repeat the same experiment for 10 times. For MARS dataset, we apply the same experiment setup in \cite{zheng2016mars}, which selects 625 identities for the training set and the remaining for testing.

The performance of all the methods are reported in Cumulated Matching Characteristics (CMC), which measures the probability that an image set in the first rank k gallery set
matches the query image set. For MARS dataset, we also report mean Average Precision (mAP) since multiple ground truth matches are available.

\subsection{Experiment Setup}
We apply two backbone networks: ResNet-50 \cite{resnet} and DenseNet-121 \cite{huang2017densely} as appearance feature extractors of our model. Both networks are pretrained on ImageNet. We extract the spatial average pooling of their last convolutional layers, which generate appearance features of 2048 and 1024 dimensions, respectively. 
During data preprocessing, input bounding box images are resized to 224 $\times$ 112. We adopt standard data augmentation steps, including horizontal flip and random crop.
In the training phase, the number of tracklets in each minibatch is 24, and 4 images are randomly selected for each tracklet.
The margin of triplet loss function is set to 0.4. 
The parameters in appearance feature extractor and identity classifier are optimized using Adaptive Moment Estimation (ADAM) algorithm \cite{adam}. 
The learning rate starts at $0.0003$, and decreases by a factor $0.1$ for each 100 epochs, until the model finishes training at 400 epochs. 
In the testing phase, all the images of a tracklet are used to extract the appearance feature set. 

\subsection{Comparison with Other Pooling Strategies}
To understand the efficacy of our proposed visual distributional representation, we compare it with other common pooling methods, including mean pooling, RNN, and temporal attention.
For a fair comparison, we adopt the same appearance feature extractors (ResNet-50 and DenseNet-121), and follow the same training scheme as our method.
Euclidean distance serves as the dissimilarity measure.

For RNN pooling, we utilize vanilla RNN architecture, and set the number of hidden units to 512.
For temporal attention pooling, we add a fully connected layer $L_a$ on top of the appearance feature extractor to calculate the attention scores. The final representation $z_i$ of the i-th image set can be obtained as followed:
\begin{equation}
\begin{split}
    z_i &= \sum_{t=1}^{n_i} a_t f_t\\
    a_t &= \frac{exp(L_a(f_t))}{\sum_{t=1}^{n_i} exp(L_a(f_t))}
\end{split}
\end{equation}
where $a_t$ is the attention score of the t-th image. 

\Cref{tab:mars1} shows the experiment results of pooling method comparison. We use a 3-dim tuple to note each possible combination of appearance feature, pooling method, and distance function.
For example, R+mean+Eu represents the model with Resnet-50 as appearance feature extractor, mean pooling strategy, and Euclidean distance as dissimilarity measure.
"KDE" and "GE" refer to two probability estimators in visual distributional representation. 
$D_S$ and $D_G$ refer to two types of Wasserstein distance approximation: Sinkhorns iteration (\Cref{WSI}) and Gaussian approximation (\Cref{WG}), respectively.

\begin{table}[]
\centering
\begin{tabular}{l|llll}
\hline
Methods        & Top-1         & Top-5         & Top-20        & mAP           \\ \hline
R + mean + Eu  & 82.9          & 93.6          & 97.1          & 76.2          \\
R + RNN + Eu   & 81.6          & 92.8          & 96.3          & 73.9          \\
R + Atten + Eu & 83.5          & 93.9          & 97.4          & 76.6          \\
R + GE + $D_G$ & 85.2          & 94.8          & 97.6          & \textbf{77.9} \\
R + KDE + $D_S$ & 84.4          & 94.6          & 97.2          & 77.5          \\ \hline
D + mean + Eu  & 82.9          & 94.2          & 97.4          & 74.5          \\
D + RNN + Eu   & 81.8          & 93.1          & 96.2          & 73.0          \\
D + Atten + Eu & 83.8          & 94.7          & 97.3          & 76.2          \\
D + GE + $D_G$  & \textbf{86.0} & \textbf{95.1} & \textbf{97.9} & 77.8          \\
D + KDE + $D_S$  & 84.8          & 94.6          & 97.4          & 77.2          \\ \hline
\end{tabular}
\caption{Comparison with common pooling strategies on MARS. R: ResNet-50, D: DenseNet121, Eu: Euclidean Distance, GE: Gaussian estimation, KDE: kernel density estimation, $D_G$: \Cref{WG}, $D_S$: \Cref{WSI} }
\label{tab:mars1}
\end{table}

From the experiment results, we can make the following observations:
(1) RNN does not outperform mean pooling baseline.
(2) Temporal attention improves performance in terms of both top-k matching accuracy and mAP. The trends are similar when we use ResNet-50 and DenseNet-121 for feature extraction.
(3) Our visual distributional representation methods outperform all the other pooling strategies. Using Gaussian approximation for Wasserstein distance estimation achieve the best performance. 
(4) In all methods, appearance feature extracted from Resnet-50 provides better performance on mAP, while that from DenseNet-121 performs better on top-k  matching accuracy.
This set of experiments demonstrates that visual distributional representation is able to preserve more information in multi-shot person re-id scenario.

\begin{figure}[t!]
  \captionsetup[subfigure]{justification=centering}
  \begin{subfigure}{0.5\textwidth}
  	\centering
    \includegraphics[width=0.7\textwidth]{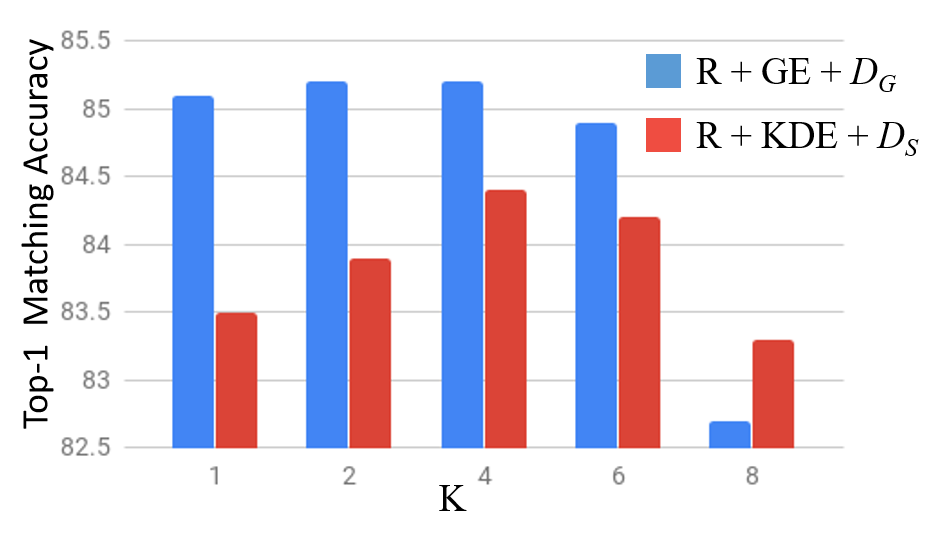}
    \caption{Experiment results on top-1 matching accuracy}
  \end{subfigure}\vspace{1em}
  \begin{subfigure}{0.5\textwidth}
    \centering
    \includegraphics[width=0.7\textwidth]{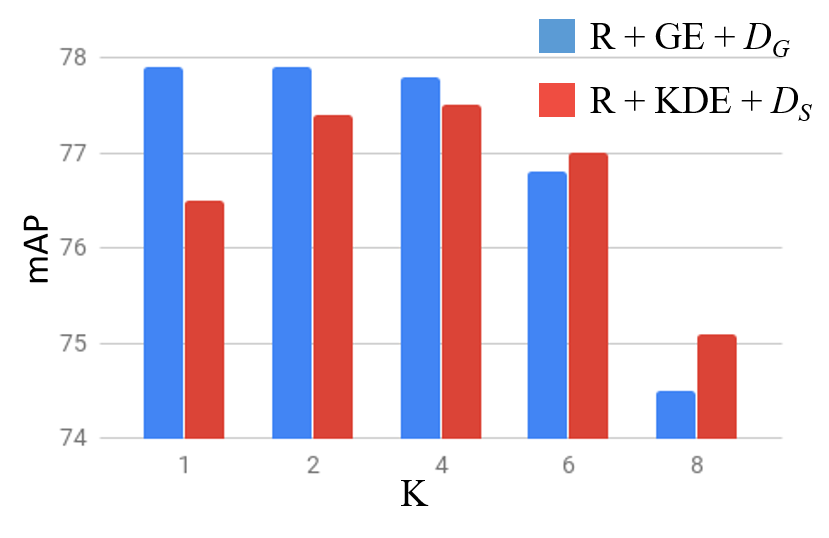}
    \caption{Experiment results on mAP}
  \end{subfigure}
  \caption{Experiment results on MARS with different window size $K$ in \Cref{MA}. ResNet-50 is utilized as appearance feature extractor.}
  \label{fig:reid3}
\end{figure}

\begin{table}[t!]
\centering
\begin{tabular}{l|l|l}
\hline
           & MARS     & iLIDS-VID \\ \hline
R+KDE+$D_S$ & Top-1   & Top-1     \\ \hline
*$\lambda$=0   & 83.9  & 77.9      \\
$\lambda$=5   & 83.9   & 78.3      \\
$\lambda$=10  & 84.4   & 79.0      \\
$\lambda$=20  & 84.3   & 79.4      \\
$\lambda$=30  & 83.8   & 78.8      \\
$\lambda$=50  & 83.7   & 78.1      \\ \hline
\end{tabular}
\caption{Top-1 accuracy on MARS and iLIDS-VID with different values of $\lambda$ in \Cref{WSI}. *The results with $\lambda=0$ is obtained by calculating the average of distance matrix $M$, instead of executing Sinkhorn iteration.}\label{tab:DS}
\end{table}

\subsection{Sensitivity Analysis}
There are two hyper-parameters in our proposed model. The first is $\lambda$ in \Cref{WSI}, which regularizes the 2-Wasserstein distance.
The second is the window size $K$ of the moving average filter in the testing phase (\Cref{MA}). 
To understand the sensitivity of these two hyper-parameters, we apply different values to our model and observe the performances.
In this experiment, we choose ResNet-50 as appearance feature extractor.

From \Cref{tab:DS}, we can tell that the re-id model receives the best performance with proper level of regularization ($\lambda=10$ or $\lambda=20$). When $\lambda=0$, the entropy regularization term in \Cref{WSI} dominates, and the value of $D_S$ becomes the average of distance matrix $M$.

\Cref{fig:reid3} illustrates the experiment results on MARS with different window size $K$. We find that the moving average filter does not affect models with Gaussian assumption (R+GE+$D_G$), since the results keep almost the same when $K=1$ . The reason may be that the Gaussian assumption has already caused some smoothing effect. However, the moving average filter is favorable for models with KDE and $D_S$. We can see that the best performance happens when $K=4$. The case of over-smoothing ($K=8$) deteriorates performances for both models.

\subsection{Comparison with State-of-the-art Methods}
We compare our proposed method to the following previous state-of-the-art methods: TDL \cite{you2016top}, CNN+XQDA \cite{zheng2016mars}, RNN \cite{mclaughlin2016recurrent}, ASTPN \cite{jointly}, SeeForest \cite{zhou2017seeforest}, LDCAF \cite{ldcaf}, QAN \cite{liu2017quality}, RQEN \cite{RQEN}, TriNet \cite{hermans2017defense}, DRA \cite{li2018diversity}, DuATM \cite{si2018dual}, and TM \cite{gao2018revisiting}. 
As suggested in previous works \cite{si2018dual,li2018diversity}, we use DenseNet-121 for MARS dataset, and ResNet-50 for iLIDS-VID dataset, respectively.

\Cref{tab:ilidsvid} summarizes the performances of all methods on iLIDS-VID dataset. We achieve competitive result comparing to state-of-the-art methods. The reason that DRA \cite{li2018diversity} performs better may be that they incorporate additional image-based re-id datasets to help with training.

The overall performances of all methods on MARS dataset are reported in \Cref{tab:mars2}. Our implementation of conventional mean pooling is already competitive. The proposed visual distributional representation can still further improve the performance without introducing additional parameters. Our method achieves the best scores in terms of both top-k matching accuracy and mAP. 

\begin{table}[t!]
\centering
\begin{tabular}{l|llll}
\hline
Methods     & Top-1         & Top-5          & Top-20        & mAP           \\ \hline
CNN+XQDA    & 68.3          & 82.6          & 89.4          & 49.3          \\
SeeForest   & 70.6          & 90.0          & 97.6          & 50.7          \\
LDCAF       & 71.8          & 86.6          & 93.1          & 56.5          \\
RQEN        & 73.7          & 84.9          & 91.6          & 51.7          \\
TriNet      & 79.8          & 91.4          & -             & 67.7          \\
*DRA         & 82.3          & -             & -             & 65.8          \\
DuATM       & 81.2          & 92.5          & -             & 67.7          \\
TM          & 83.3          & 93.8          & 97.4          & 76.7          \\ \hline
Ours (D+mean+Eu) & 82.9          & 94.2          & 97.4          & 74.5          \\
Ours (D+KDE+$D_S$) & 84.8          & 94.6          & 97.4          & 77.2          \\
Ours (D+GE+$D_G$) & \textbf{86.0} & \textbf{95.1} & \textbf{97.9} & \textbf{77.8} \\ \hline
\end{tabular}
\caption{Comparison with state-of-the-art methods on MARS. *: additional training datasets used.}\label{tab:mars2}
\end{table}

\begin{table}[t!]
\centering
\begin{tabular}{l|lll}
\hline
Methods           & Top-1         & Top-5         & Top-20        \\ \hline
TDL               & 56.3          & 87.6          & 98.3          \\
RNN               & 58            & 84            & 96            \\
ASTPN             & 62            & 86            & 98            \\
CNN+XQDA          & 53.0          & 81.4          & 95.1          \\
SeeForest         & 55.2          & 86.5          & 97.0          \\
QAN               & 68.0          & 86.8          & 97.4          \\
RQEN              & 76.1          & 92.9          & \textbf{99.3} \\
*DRA               & \textbf{80.2} & -             & -             \\ \hline
Ours (R+mean+Eu)  & 77.0          & 93.3          & 98.4          \\
Ours (R+GE+$D_G$)  & 77.2          & 93.3          & 98.4          \\
Ours (R+KDE+$D_S$) & 79.4          & \textbf{93.8} & 98.7          \\ \hline
\end{tabular}
\caption{Comparison to state-of-the-art methods on iLIDS-VID. *: additional training datasets used.}\label{tab:ilidsvid}
\end{table}

\subsection{Visualization}
\begin{figure}
  \centering
    \includegraphics[width=0.47\textwidth]{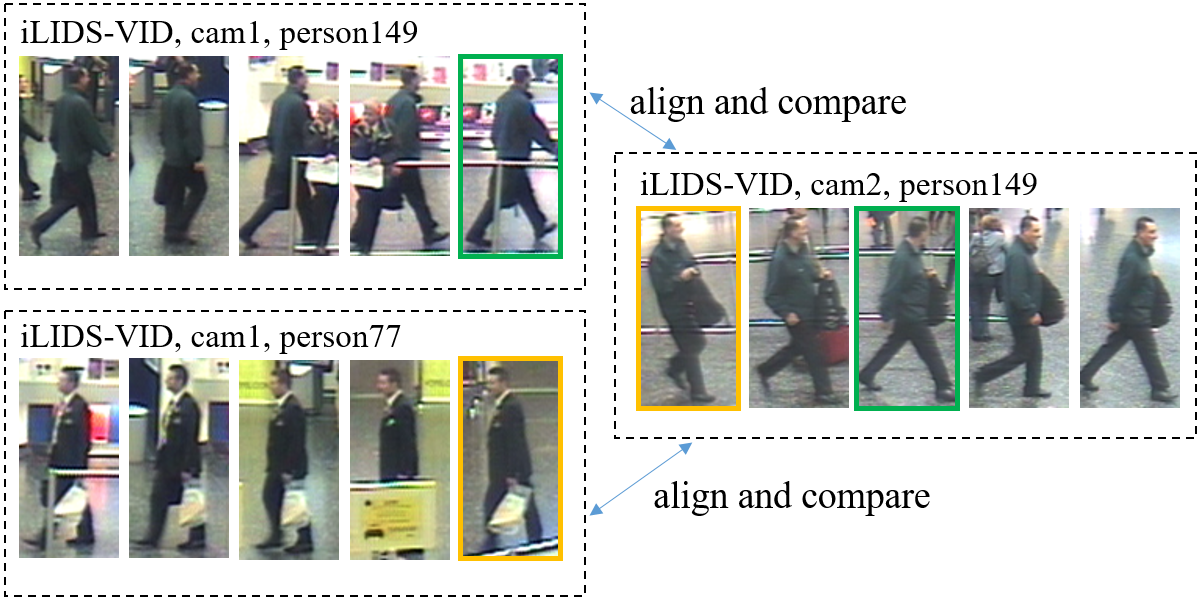}
  \caption{Visualization of the alignment automatically discovered by our proposed method. Image pairs assigned with highest joint probability are marked by bounding boxes in the same color.}
  \label{fig:reid_vis}
\end{figure}

We show two examples of alignment between person tracklets in \Cref{fig:reid_vis}.
Three person tracklets are picked from iLID-VIDS dataset.
Two of them belong to the same identity, and the rest one is different.
From \Cref{fig:reid_vis}, one can see that our proposed method attends to different images of a tracklet while comparing to different candidates. Please note that the images without color bounding boxes are manually selected to indicate the appearance of the whole tracklet.

\section{Conclusion}
In this paper, we propose to learn visual distributional representation with Wasserstein distance, and conduct set-to-set comparison for multi-shot person re-identification task.
Our approach can effectively discover a proper alignment between input query and gallery examples without additional parameters.
Experiment results show that our proposed method outperforms several common feature aggregation strategies, and achieve competitive/superior performance comparing to previous state-of-the-art approaches.

The proposed visual distributional representation provides a general strategy to summarize a set of images, and can be easily plugged into an end-to-end learning architecture.
Our future work will focus on the exploration of downstream tasks other than multi-shot person re-id.

\bibliographystyle{aaai}
\bibliography{strings}
\end{document}